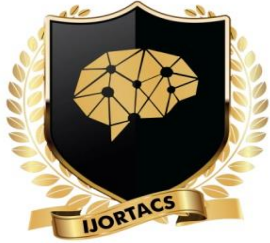

## DEVELOPMENT OF AN INTELLIGENT SYSTEM FOR THE DETECTON OF CORONA VIRUS USING ARTIFICIAL NEURAL NETWORK

[1]Nwafor Emmanuel O., [2]Ngozi Maryrose Umeh., [3]Ikechukwu Ekene Onyenwe

[1,2,3]Nnamdi Azikiwe University, Awka, Anambra State.

[1]nwaforemmanuel041@gmail.com,

### Abstract

*This paper presents the development of an intelligent system for the detection of coronavirus using artificial neural network. This was done after series of literature review which indicated that high fever accounts for 87.9% of the COVID-19 symptoms. 683 temperature data of COVID-19 patients at $\geq 38C^o$ were collected from Colliery hospital Enugu, Nigeria and used to train an artificial neural network detective model for the detection of COVID-19. The reference model generated was used converted into Verilog codes using Hardware Description Language (HDL) and then burn into a Field Programming Gate Array (FPGA) controller using FPGA tool in Matlab. The performance of the model when evaluated using confusion matrix, regression and means square error (MSE) showed that the regression value is 0.967; the accuracy is 97% and then MSE is 0.00100Mu. These results all implied that the new detection system for is reliable and very effective for the detection of COVID-19.*

### I.    INTRODUCITON

Coronavirus (COVID-19) is a disease that has shocked the world and has impacted every sector in it. Fortunately the mankind is gradually recovering from the setbacks caused by the epidermis in all walks of life; however a lot still have to be done to facilitate the control measures. According to [1], COVID-19 is a colony of virus that causes illness ranging from the common cold to more severe symptoms such as severe acute respiratory syndrome and even death. This disease remain a very big threat to the existence of man till date, despite the hug efforts put in place so far to control it.

Since the inception of this epidermis, one widely adopted approach to test and detect it is using thermal sensor. This approach was employed based on the fact that high fever remain the most common symptoms of COVID-19 patients with over 87.9% of the patients recording it as first sign after infection[2]. However this thermal sensors used all over the world today were not developed for corona virus fever; but rather for normal high fever which has been in existence for many decades before the evolution COVID-19. Secondly the safe distance of at least 1.5m recommended by [3] further increases the limitations of this sensor as they were not designed to operate at this safe distance from the patients, hence their sensitive and accuracy is poor, making them no reliable for the detection of COVID-19.

[4] opined that due to this safe distances specification, the existing thermal sensor only correctly detect one out of five COVID-19 patients. This result is very poor and the reason is clear, "the sensor was not made for COVID-19" but was employed as a panic measure to detect the virus and help control the transmission rate.

The sensor later gained more application due to the fact that other means of testing COVID-19 such as serology, imaging, blood test, Reverse transcription polymerase chain reaction (RTPCR), antigen, Isothermal amplification assays e.tc all takes between 9 to 21 days to produce results [5; 6; 7]. Hence this sensor is the only alternative for fast COVID-19 detection based on high fever. However due to the limitation of the thermal sensor, the study propose to improve the performance using artificial neural network (ANN).

ANN is a branch of machine leaning algorithm which has the ability to learn and make accurate decisions using training dataset [8]. This system will be developed, trained with temperature data collected from a COVID-19 hospital and deployed to improve the performance of the conventional thermal sensor and make it intelligence. This when achieved will make the conventional sensor more intelligence, accurate and reliable for real time detection of COVID-19.

## II.    LITERATURE REVIEW

In [9] they presented a research work on the detection and analysis of coronavirus disease. The work was done using adopting the logistic model, Bertalanffy model and Gompertz model for the detection design. This detection was done based on the 4,879 confirmed cases in China as at 28 February, 2020. [10] Presented a wok on the modified SEIR and A.I detection of the epidemic trend of the COVID-19 in china. The work was developed using the data collected from the Wuhan hospital under the guidance of the china public health intervention agency. The work was able to employed the unsupervised machine learning technique (support vector machine) for the training and diagnostic detection of the covid-19 patients. They induced that the use of other A.I techniques like the artificial neural network will produce a better result, superseding theirs with detection accuracy of 89%. In [11] a research work based on automatic organ segmentation of CT scan based on super pixel and convolutional neural network was presented. The work was able to analyzed scanned images of infected individuals with certain viral infections like SARS (also a COVID-19 symptom) using the image processing techniques and then trained the result to detect further query scanned input. This work is applicable for radiography and can also be applied to treat or detect COVID-19 cased associated with breathing issues. [12] Presented a research using mathematical model for the detection of COVID-19 epidermis. The model was able to provide background on COVID-19, symptoms and challenges and recommends testing as a measure to control the spread. [13] Presented a research on the early estimation of the epidemiological parameters and epidemic detection of the novel coronavirus. This work shows mathematically using process models and data collected from the Wuhan general hospital to detect the spreading rate of the virus. The research shows how fast the virus can spread from a host if not controlled. They induced that community transmission can be experience later within the Wuhan city if care is not taken.

## III. METHODS

The methods used for development of the proposed system are temperature sensor, data collection, feature extraction, artificial neural network, training, integration and classification.

**Temperature Sensor:** This is the conventional system employed all over the world today for the detection of the COVID-19 through high fever. The system was developed using thermal sensor which detects latent heat from the body and then record the result, however the efficiency in the result was limited due to the COVID-19 safe distance. This limitation was addressed in this research using data collection and artificial neural network.

**Data collection:** data was collected for the study from the Enugu State Colliery Hospital Nigeria (Only authorized hospital to treat coronavirus in the state as at the time of the research). The data was requested for this research purpose only and was used to train the machine learning algorithm proposed. The collected data contains attributes of confirmed COVID-19 patient's body temperature with a sample size of 693 samples at temperature of $\geq 38C^o$.

**Feature extraction:** this process involves the extraction of the temperature data from the training dataset into compact feature vectors which was feed forward to the neural network for training.

**Artificial Neural Network:** this is a brain of machine learning biologically inspired by the human brain with the capacity to learn and make accurate decisions using its neurons, bias and activation function with the help of a training algorithm. The ANN was used to train the temperature data collected and then generated a reference model which was incorporated on the conventional thermal sensor to improve the performance.

**Training:** this process allows the neural network to learn the COVID-19 temperature features from the training dataset via system identification and then generate a reference COVID-19 model which will be used as a base for time series detection of COVID-19.

**Integration:** this process involves the deployment of the reference COVID-19 model on the conventional temperature sensor to improve the performance. This was done marrying the reference ANN algorithm developed with the thermal sensor plant to make it intelligence as the new improved COVID-19 sensor. This was done using Simulink.

**Classification:** this process involves the ability of the new sensor developed to make accurate decision. This was involves collecting time series data of suspected patients and then comparing the feature with the pattern of the reference model to make decision if the patient have COVID-19 or not.

## IV.    SYSTEM DESIGN OF THE NEW SENSOR

The new system was designed using artificial neural network to identify the temperature features extracted from the training dataset as a nonlinear autoregressive model (NARX), and then used training back propagation algorithm to train the model to achieve a reference COVID-19 model which was used to improve the conventional temperature sensor. The model of the conventional temperature sensor was presented in figure 3.1;

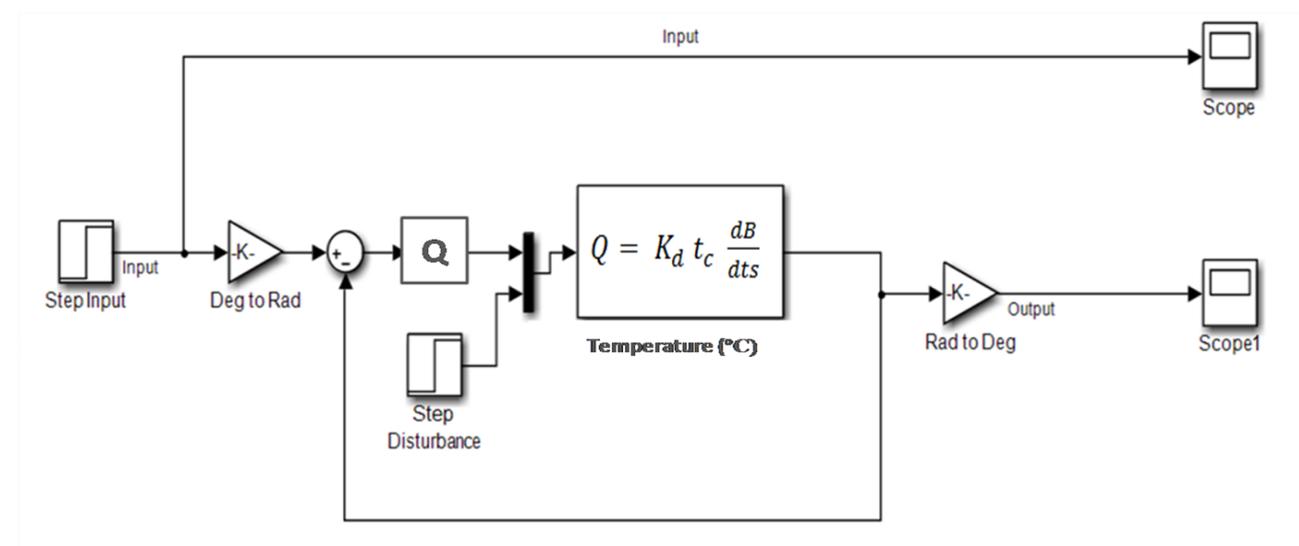

Figure 1: Model of the conventional thermal sensor

The figure 1 presented the model of the conventional temperature sensor; where Q is the net heat flow into port A; $K_d$ is the dissipation factor parameter value; $t_c$ is the thermal time constant parameter value; dB/dt is the rate of change of the temperature and s is distance.

**Artificial neural network Modeling Diagram**

The modeling diagram of the artificial neural network was presented using the activity model in figure 2. The model was developed using the input neurons which uses bias and weight to identify data feed to the neurons and sum them up to activate using activation function in [14]. The activated output is trained to learn the patterns and generate a reference model. The model of the neural network when trained with the COVID-19 data presented in figure 3 using logical flow chart; while the training parameters are presented in table 1.

**Table 1; ANN training parameters**

| Parameters | Values | Parameters | Values |
|---|---|---|---|
| Number of epoch value | 11 | Number of non hidden layers | 128 |
| Size of hidden layers | 824 | Maximum interval per sec | 2 |
| No. delayed reference input | 9 | No. delayed controller output | 1 |
| Maximum plant output | 2 | No. delayed plant output | 2 |
| Number of feature input | 696 | Minimum reference value | -0.05 |

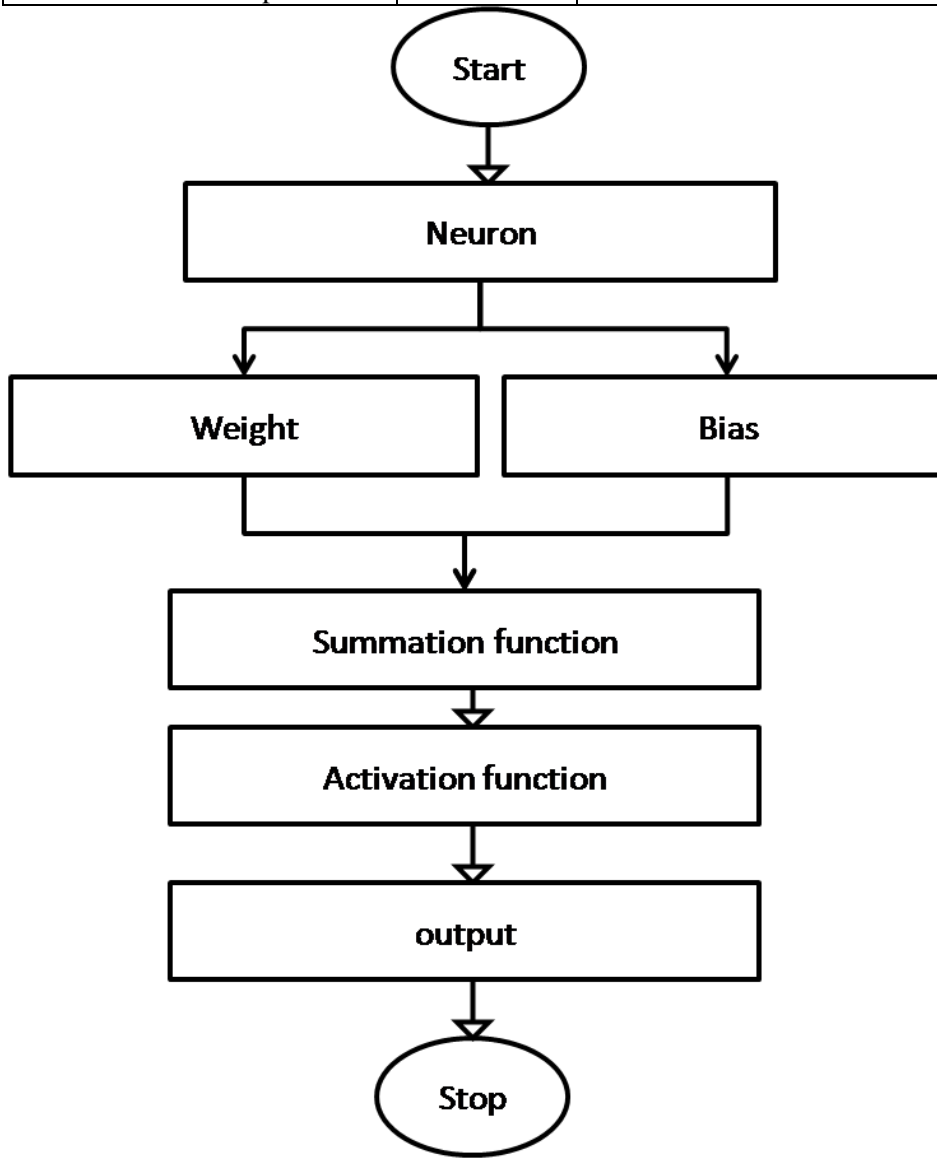

Figure 2: Flow chart of the ANN architecture

The figure 2 presented the workflow of the neural network model which was used to train the data collected. The neural network is made up of neurons, bias and activation functions. The activation function used is the tansig activation function whose work is to trigger the neurons to give output when data was loaded for training. The flow chart of the training process which generated the reference model used for the detection of covid-19 was presented in the figure 3;

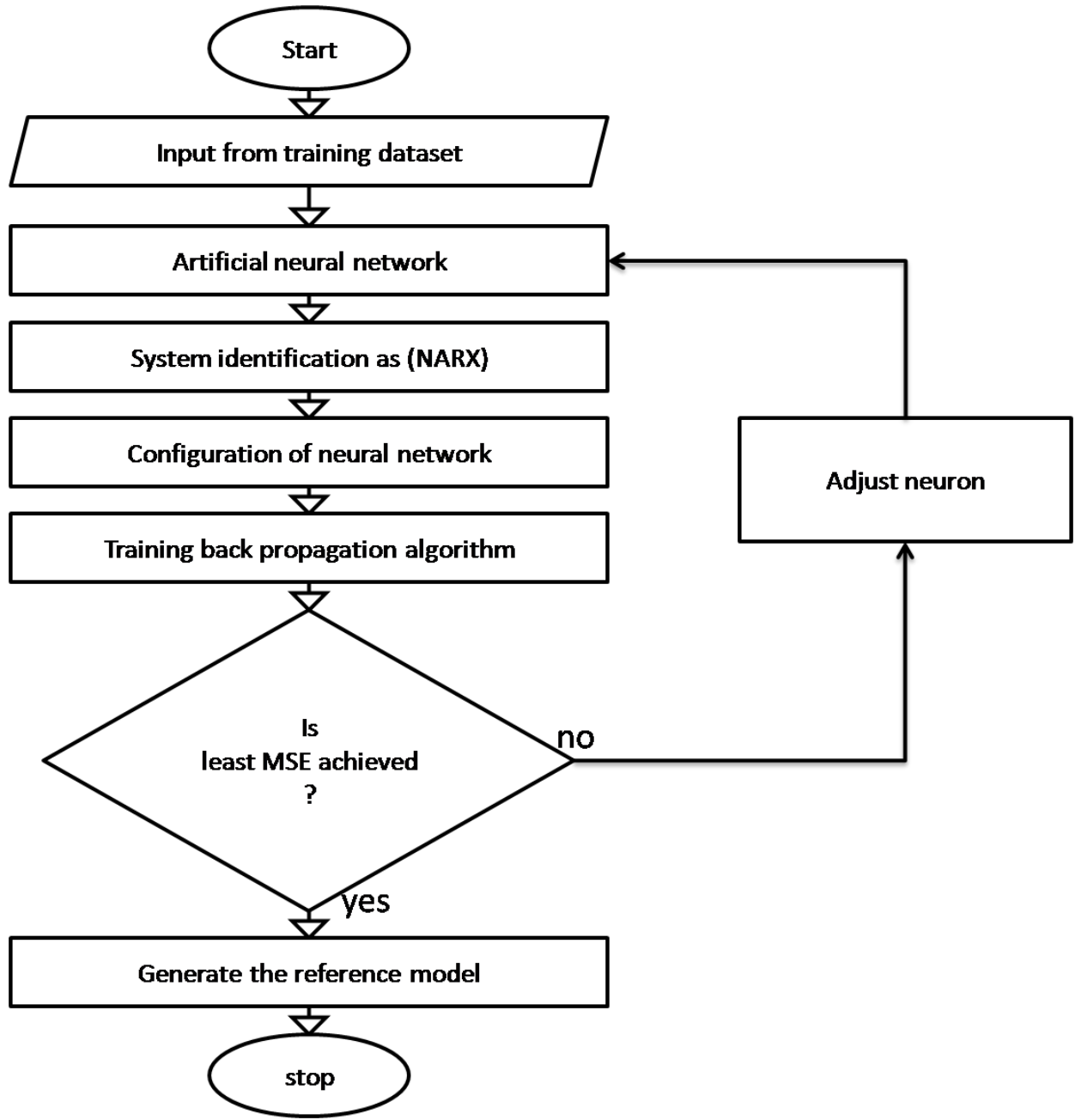

Figure 3: Flow chart of the generated model for Covid-19 detection

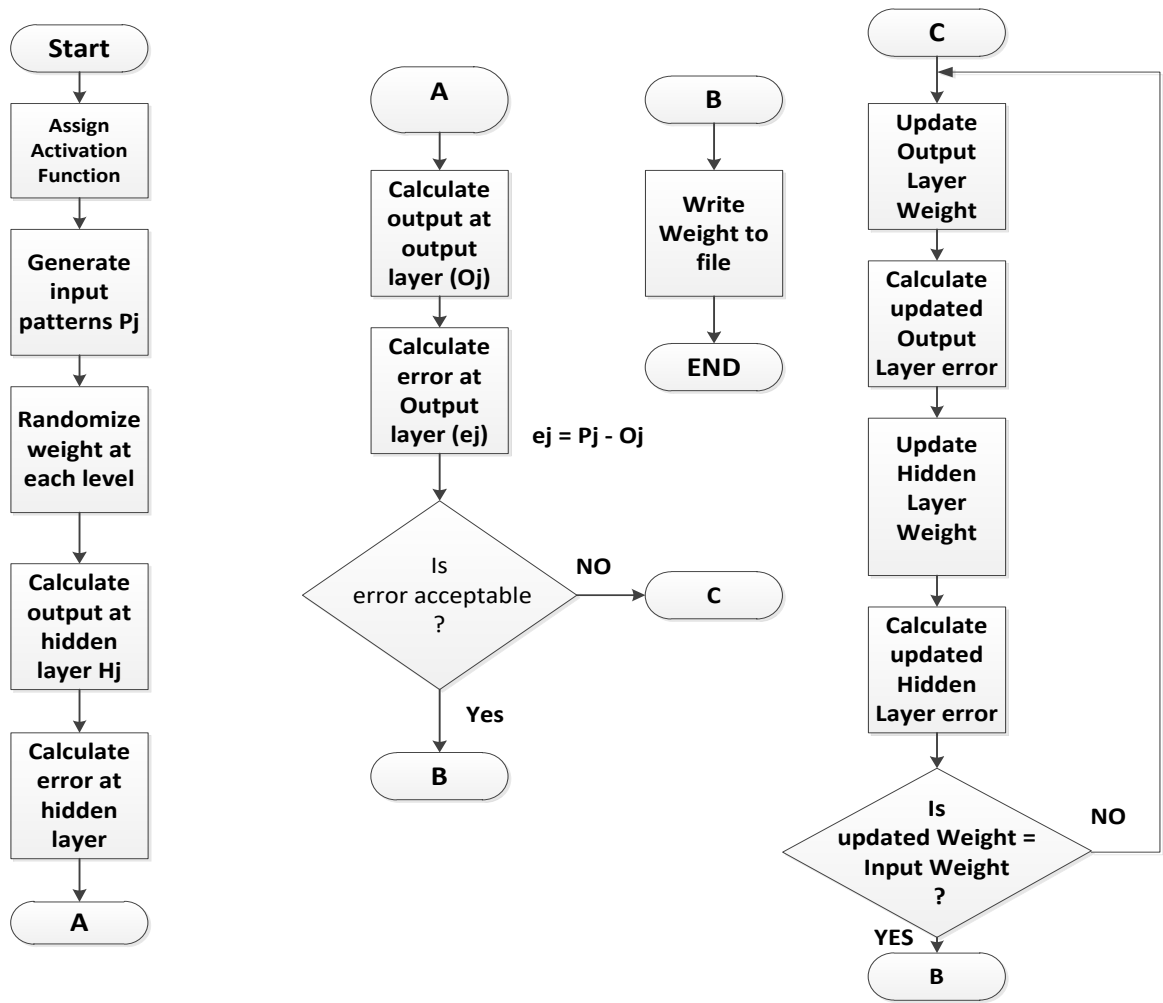

Figure 4: flow chart for back-propagation algorithm for the ANN training

**Detective model**

The neural network detects the network response over the specified input vector (testing dataset). The detector is a numerical optimization programs which determine the feature vectors that minimizes the training performance criterion over the specified horizon as shown below;

$$J = \sum_{j=N1}^{N2}(y_r(t+j) - y_m(t+j))^\wedge 2 + p\sum_{j=1}^{Nu}(u'(t+j-1) - u'(t+j-2))^\wedge 2 \qquad (1)$$

Where $N_1$, $N_2$, and $N_u$ define the horizons over which the training error and the detection features are evaluated. The $u'$ variable is the tentative feature vectors from the test dataset, $y_r$ is the desired response, and $y_m$ is the reference model, t is time. The p value determines the contribution that the sum of the squares of the control increments has on the performance index. The model consists of the neural network training model and the optimization block. The optimization block determines the values of $u'$ that minimize J, and then the optimal u is input to the network.

## V.    IMPLEMENTATION

To implement the system, neural network toolbox, statistics and machine learning toolbox, the modeling diagrams and the mathematical models developed were all used to implement the system using Simulink as show in the figure 5; The toolbox before training automatically splits the data into multiset of training, test and validation sets in the ratio of 70:15:15 respectively

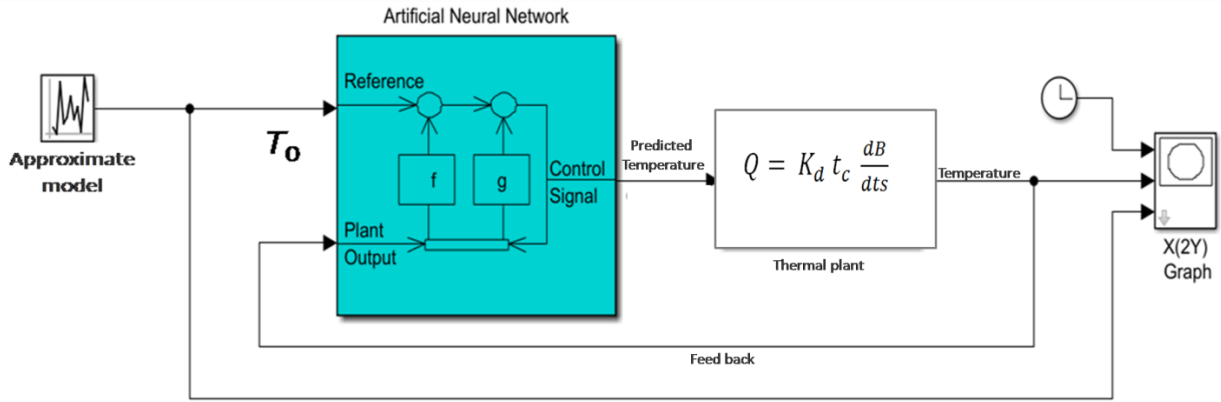

Figure 5: Simulink model of the Neuro COVID-19 Sensor

The figure 5 presented the Simulink model of the control system and how it was attacked to the temperature sensing element to feedback the sense input to the controller system for training with the reference model and then detect COVID-19. To convert into hardware, the reference model generated was used converted into Verilog codes using Hardware Description Language (HDL) and then burn into a Field Programming Gate Array (FPGA) controller using FPGA tool.

**Performance metrics**

The performance of the neural network will be evaluated using the confusion matrix, training accuracy (Acc), true positive rate and false positive rate. These are measuring tools are guided by the respective models below;

$$Accuracy\ (ACC) = \frac{TP+TN}{TP+TN+FP+FN} \tag{2}$$

$$True\ Positive\ Rate\ (TPR) = \frac{TP}{TP+FN} \tag{3}$$

$$False\ positive\ Rate\ (FPR) = \frac{TN}{TN+FP} \tag{4}$$

Where TP = true positive, TN = true negative, FN =false positive, FN = false negative

## VI.    RESULTS AND DISCUSSION

The system is implemented using the models developed in the system design section, alongside neural network toolbox, statistics and machine leaning toolbox, control system toolbox and Mathlab. The model was simulated using the training parameters in table 1, and the training result was generated using neural network training toolbox. The result presents confusion matrix, regression and mean square error performance of the system.

The figure 6 presented the confusion matrix which was used to determine the COVID-19 detection accuracy when deployed for testing. The result is presented as;

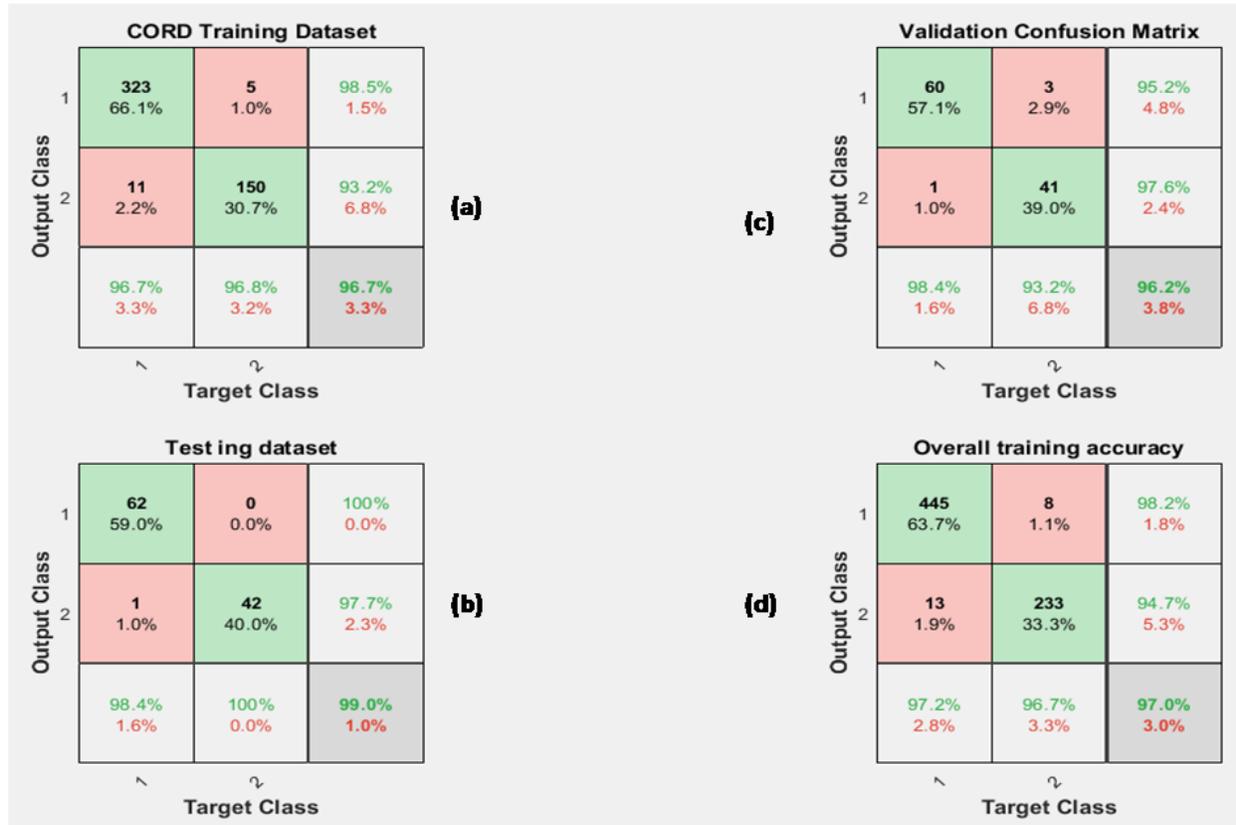

Figure 6: confusion matrix performance

From the confusion matrix, the training performance of the feature vectors extracted from the multi set is evaluated. The training dataset (a) correctly classified COVID-19 with 96.7% accuracy. On the testing dataset (b) also trained, the rate of correct classification is 99%. The training performance on the multiset is summarized using overall training result (d) which calculated the average training performance of the multiset with correct classification accuracy of 97%. The implication of this confusion matrix result shows that the detection rate of the new system will 97% correct when trained with new testing dataset. This confusion matrix is further analyzed using the validation curve considering the sensitivity and specificity of the training process as shown below.

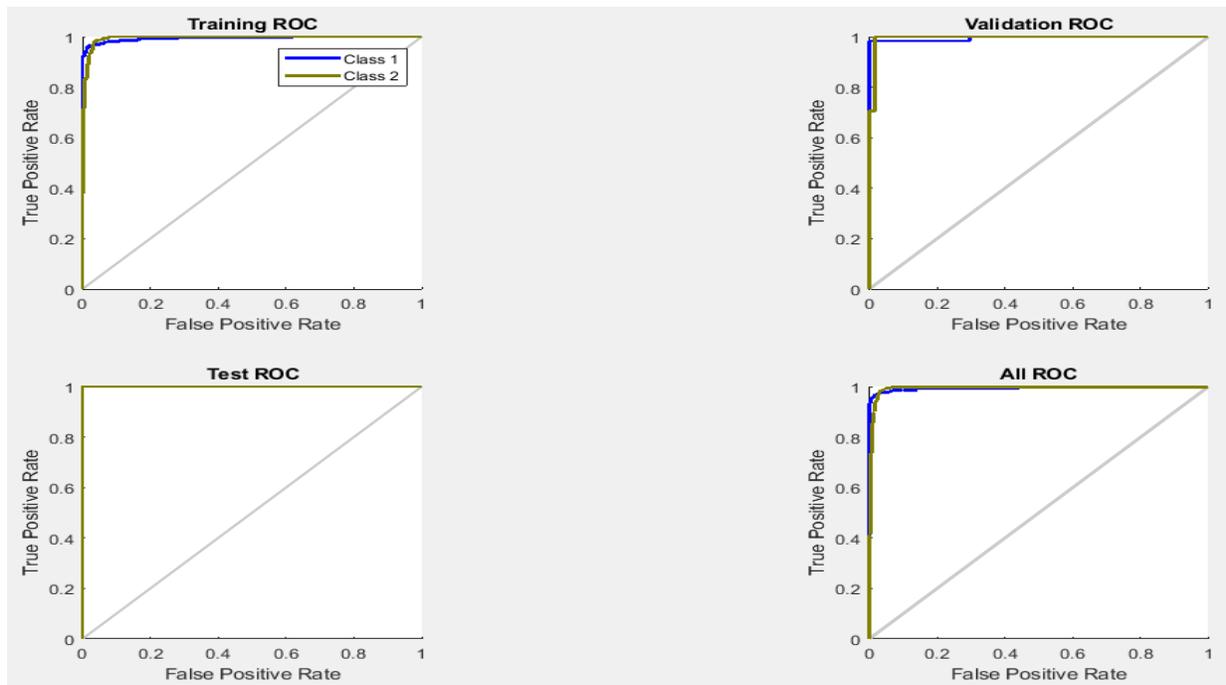

Figure 7:  receiver operator characteristic curve

The figure 7; expresses the receiver operator characteristics of the training process. The result shows that a good regression (R) value for the overall system performance was achieved with (R = > 0.96). The result is made more précised using the table 2;

**Table 2: Performance of the ROC curve**

| Multiset | False positive rate% | True positive rate% |
|---|---|---|
| Training | 0.033 | 0.967 |
| Testing | 0.010 | 0.990 |
| Validation | 0.038 | 0.962 |
| Overall training performance | 0.030 | 0.970 |

From the table 2; the regression (R) performance of the neural network as displayed in figure 7 is presented. The aim is to achieve a regression value approximately or equal to 1. Thus from the result recorded, the overall regression performance is (R=0.97). The implication of these results shows that performance of the ANN is précised since the regression value is approximately = 1. This point which the ANN performs best and achieved this result is called the best validation performance. At this point the training process stops automatically, this process is monitored at each state of the training using incremental epoch value. At the point the training performs best, the epoch value is recorded and the training stops. This is analyzed using the mean square error result in figure 8;

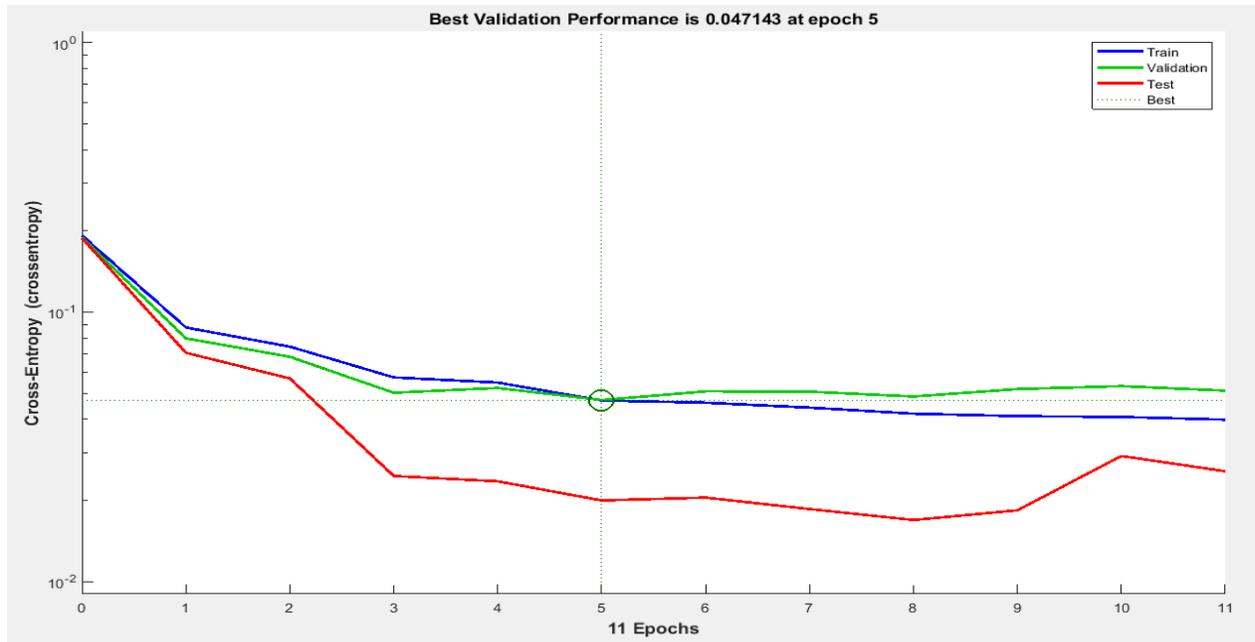

Figure 8: best validation performance

From the result in figure (8), it was observed that a mean square error (MSE) performance of 0.00100Mu was achieved at epoch 5 and validation value of 0.047143. The implication of this result showed that the margin for error in the new system is almost none which is acceptable and hence justified how effective and reliable the new system is when deployed for the detection of COVID-19.

## VI. CONCLUSISON

This work has successfully presented an intelligent COVID-19 detection and detection system with high performance and accuracy. This was achieved improving the conventional temperature sensor with artificial neural network using temperature data collected from COVID-19 patients. The result when tested showed that the new system is reliable, high efficient with a detection accuracy of 97%. This when deployed for general use will improve the rate of detecting COVID-19 in outdoor and indoor environments and help control the spread of the epidermis.

### VII.    CONTRI BUTION TO KNOWLEDGE
a)    A neuro COVID-19 detection system was developed
b)    A system which can detect COVID-19 with regression of 0.970 was achieved
c)    A near real time system for the detection of COVID-19 was presented

## VIII.   Recommendation

a) Due to the fact that some patients do no show symptoms, other testing technique can be used to detect COVID-19

b) The system despite the success in detecting COVID-19, there is need for other test to confirm the virus in patients.